\documentclass[lettersize,journal]{IEEEtran}
\usepackage{amsmath,amsfonts}
\usepackage{algorithmic}
\usepackage{algorithm}
\usepackage{array}
\usepackage[caption=false,font=normalsize,labelfont=sf,textfont=sf]{subfig}
\usepackage{textcomp}
\usepackage{stfloats}
\usepackage{url}
\usepackage{verbatim}
\usepackage{graphicx}
\usepackage{cite}
\usepackage{makecell}
\usepackage{multirow}
\usepackage{multicol}
\usepackage{booktabs}
\usepackage{bigstrut}

\usepackage{xspace}
\usepackage{float}
\usepackage{pifont}
\usepackage{amssymb}
\def\onedot{.\xspace}
\def\eg{\emph{e.g}\onedot} 

\def\ie{\emph{i.e}\onedot}

\begin{document}

\title{Toward Real-world Single Image Deraining: A New Benchmark and Beyond}

\author{Wei Li$^{*}$, Qiming Zhang$^{*}$,~\IEEEmembership{Student Member,~IEEE}, Jing Zhang,~\IEEEmembership{Member,~IEEE}, Zhen Huang,  Xinmei Tian,~\IEEEmembership{Member,~IEEE}, and Dacheng Tao,~\IEEEmembership{Fellow,~IEEE}
        % <-this % stops a space
}


\maketitle
\def\thefootnote{*}\footnotetext{Equal contribution.}
\begin{abstract}
Single image deraining (SID) in real scenarios attracts increasing attention in recent years. Due to the difficulty in obtaining real-world rainy/clean image pairs, previous real datasets suffer from low-resolution images, homogeneous rain streaks, limited background variation, and even misalignment of image pairs, resulting in incomprehensive evaluation of SID methods. To address these issues, we establish a new high-quality dataset named RealRain-1k, consisting of $1,120$ high-resolution paired clean and rainy images with low- and high-density rain streaks, respectively. Images in RealRain-1k are automatically generated from a large number of real-world rainy video clips through a simple yet effective rain density-controllable filtering method, and have good properties of high image resolution, background diversity, rain streaks variety, and strict spatial alignment. RealRain-1k also provides abundant rain streak layers as a byproduct, enabling us to build a large-scale synthetic dataset named SynRain-13k by pasting the rain streak layers on abundant natural images. Based on them and existing datasets, we benchmark more than 10 representative SID methods on three tracks: (1) fully supervised learning on RealRain-1k, (2) domain generalization to real datasets, and (3) syn-to-real transfer learning. The experimental results (1) show the difference of representative methods in image restoration performance and model complexity, (2) validate the significance of the proposed datasets for model generalization, and (3) provide useful insights on the superiority of learning from diverse domains and shed lights on the future research on real-world SID. The datasets have been released at https://github.com/hiker-lw/RealRain-1k.
\end{abstract}


\section{Introduction}
\begin{figure*}[!t]
    \centering
    \includegraphics[width=\linewidth]{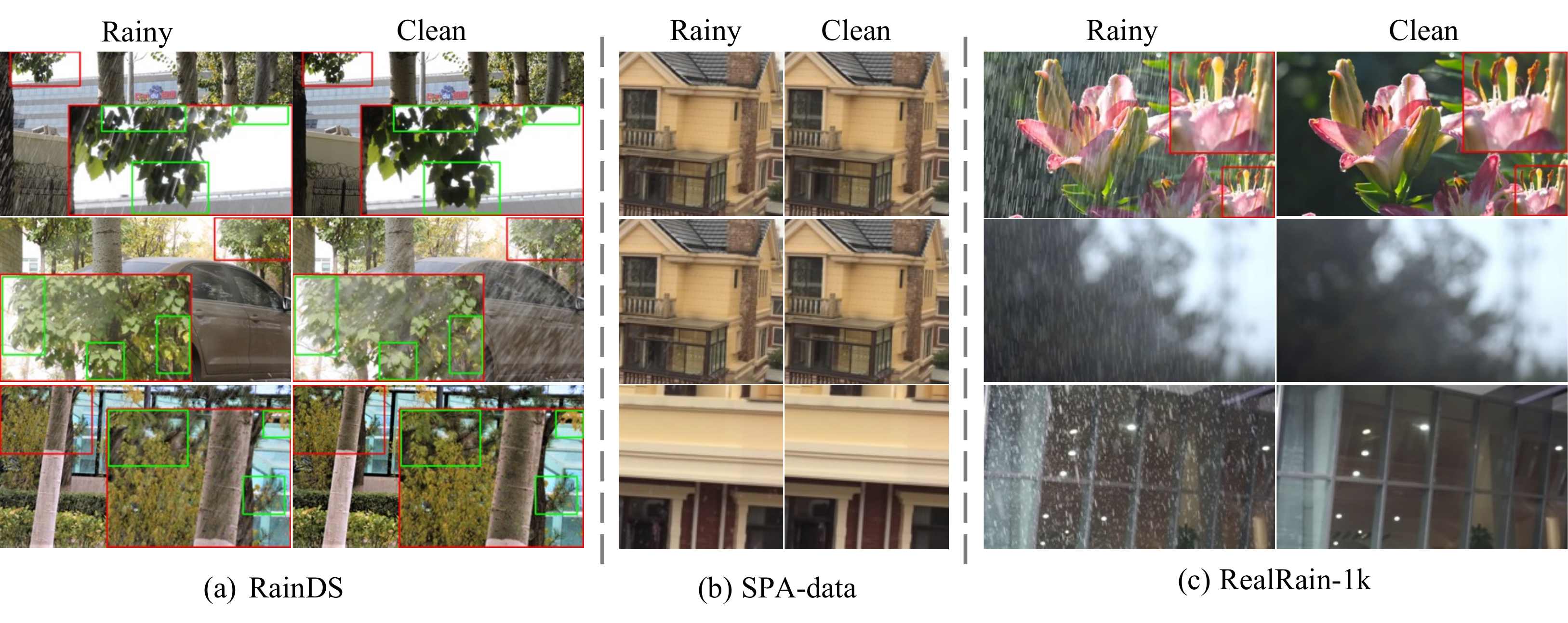}
    \caption{A glance at different real rainy datasets. (a) RainDS. (b) SPA-data. (c) our RealRain-1k.
    % The left column in each group refers to rainy images, while the right column are the paired clean ones.
    % Close-up views of some patches are enclosed by red rectangles.
    The image patches in red rectangles are enlarged for better visual comparison.
    Some misaligned regions in RainDS are indicated by green rectangles.
    The diversity of rain streaks and background appearance in SPA-data is relatively limited.}
    \label{fig:dataset_glance}
    % \vspace{-2 mm}
\end{figure*}
\IEEEPARstart{I}{mages} captured in rainy days suffer from noticeable degradation of visual quality, thereby affecting both human visual perception and practical applications of advanced computer vision systems such as autonomous driving \cite{wang2021context,zhang2020empowering}. The aim of single image deraining (SID) is to restore the clean and rain-free background images from the rainy ones. Benefitting from the rapid development of deep convolutional neural networks (CNNs) \cite{lecun2015deep}, recent years have witnessed an explosive spread of using CNN models to perform SID \cite{zamir2021restormer,zamir2021multi,yi2021structure,chen2021hinet,deng2020detail,wang2019spatial,li2018recurrent,fu2017removing}.

Recently, many research efforts have been made to solve SID from synthetic images to the real-world rainy images, where two real datasets are proposed, \ie, SPA-data~\cite{wang2019spatial} and RainDS~\cite{quan2021removing}. While they contribute to advancing the research for real-world SID, the difficulty in collecting high-quality rainy and clean pairs in the real world leads to various issues, including low resolution ($256\times256$), homogeneous rain streaks, limited background variation (Figure~\ref{fig:dataset_glance} (b)), and even spatial misalignment (Figure~\ref{fig:dataset_glance} (a)), limiting the generalization ability of SID models trained on them in dealing with various real-world rainy images and implying limited practical significance of them.

As a result, there are several important research questions that have not been well studied or even remain unexplored: 1) How the existing SID methods perform in complex real-world scenarios, where the rainy images are captured at high resolutions and have diverse rain streaks and backgrounds? 2) How about the quantitative performance of existing methods when generalizing to an unseen real scene? 3) How to well utilize existing large-scale synthetic datasets to obtain a better performance on real rainy images? 

To answer these questions, it is necessary to establish a new high-quality real-world dataset and pave the way for future research in this field. In this paper, we make an attempt towards this goal by proposing two datasets, RealRain-1k and SynRain-13k. Specifically, we first collect $1,120$ high-resolution rainy videos, covering different scenes with diverse rain streaks and abundant backgrounds. Then, we use a simple yet effective method to control the rain density in our generated images and carefully check their spatial alignment. In this way, we obtain $1,120$ high-quality image triplets (of clean, light rain, heavy rain), resulting in two versions of our RealRain-1k dataset, \ie, RealRain-1k-L and RealRain-1k-H, respectively.

Besides, RealRain-1k also provides abundant real rain streak layers as a byproduct by reversing the linear superimposition model on clean and rainy images. We thus construct a large-scale synthetic dataset named SynRain-13k by pasting the rain streak layers on abundant natural images. It has a smaller domain gap to real rainy images compared with previous synthetic datasets, thereby facilitating training better SID models.

Based on RealRain-1k and SynRain-13k, we benchmark more than ten SID models on three tracks: 1) SL track, where these models are trained and evaluated on RealRain-1k in a supervised learning manner. We compare their performance in terms of restoration quality, model complexity, inference throughput, and convergence speed; 2) DG track, where these models are trained on different datasets and tested on unseen real datasets to investigate their syn-to-real and real-to-real domain generalization ability as well as the value of synthetic datasets; 3) TL track, where SID models are pre-trained on synthetic datasets, further fine-tuned and tested on real datasets to investigate the benefit of transfer learning. Based on the experiment results, we gain some useful insights on the limitations of existing SID methods and datasets and discuss future research directions.

The main contribution of the paper is two-fold. (1) We establish a high-quality real dataset RealRain-1k for SID study, which consists of 1,120 high-resolution clean and rainy image pairs with diverse rain streaks, abundant backgrounds, and strict spatial alignment. We also generate a large-scale synthetic dataset of 13k image pairs, \ie, SynRain-13k, by leveraging the diverse real rain layers from RealRain-1k. (2) Based on them, we comprehensively benchmark more than ten representative SID methods on three tracks including SL, DG and TL. The experiment results validate the value of our datasets and provide useful insights on the limitations of existing SID methods. 
\section{Related work}
\label{gen_inst}
\subsection{Single image deraining}
\textbf{Traditional methods} Generally, SID methods can be categorized as prior-based methods and data-driven deep learning ones. Traditional methods mainly use hand-crafted features and priors to describe the features of rain streaks. For example, Kang et al. \cite{kang2011automatic} propose to first decompose an image into low and high frequency parts and remove rain streaks in the high frequency layer via dictionary learning. Luo et al. \cite{luo2015removing} present a discriminative sparse coding method for separating rain streaks from the background. Li et al. \cite{li2016rain} use Gaussian mixture models (GMM) as the prior to model rain streaks and background separately for rain removal. Zhu et al. \cite{zhu2017joint} utilize layer-specific priors to discriminate rainy regions. Albeit achieving good performance on certain scenarios, the prior-based methods rely on specific priors, which may not applicable to many real scenes where the rain streaks are complex and diverse. As a result, they have a limited generalization ability. 

\textbf{CNN-based methods} Recently, many deep-learning based approaches~\cite{yasarla2019uncertainty,jiang2020multi,wang2019spatial,yang2017deep,li2018recurrent,fu2017removing,deng2020detail,ren2019progressive,wang2020model,yi2021structure,jiang2020decomposition,ahn2021eagnet,zhu2020learning,cai2022multi}, have been proposed for rain streak removal and made significant progress in this area. For example, Fu et al. first adopt a shallow CNN \cite{fu2017clearing} and then a deeper ResNet \cite{fu2017removing} to remove rain streaks. 
%Yang et al. \cite{yang2017deep} proposed a recurrent deep network for joint rain detection and removal to progressively remove rain streaks. 
Zhang et al. \cite{zhang2017convolutional} take the rain density into account and present a multi-task CNN for joint rain density estimation and deraining. %Li et al. \cite{li2018recurrent} proposed a recurrent squeeze-and-excitation context aggregation network to utilize the useful contextual information. 
Ren et al. \cite{ren2019progressive} provided a simpler baseline deraining network by considering network architecture, input and output, and loss functions. 
%Deng et al. \cite{deng2020detail} introduced two parallel subnetworks that synergize to derain and recover lost details during deraining. 
Jiang et al. \cite{jiang2020multi} proposed a multi-scale coarse-to-fine progressive fusion network to remove rain streaks. 
%Wang et al. \cite{wang2020model} introduced a rain convolutional dictionary network and utilized the proximal gradient descent to optimize the deraining network. 
%Yi et al. \cite{yi2021structure} proposed a structure preserving deraining network to directly generates high-quality rain-free images with clear and accurate structures under the guidance of residue channel prior. 

\textbf{Transformer-based methods} Vision transformer \cite{vaswani2017attention,dosovitskiy2020image} has recently been introduced in image restoration tasks and obtains significant performance due to its powerful modeling ability~\cite{liu2022tape,cai2022hipa,yang2020learning,kumar2020colorization,ji2021u2}. Chen et al. \cite{chen2021pre} first applies a standard transformer block equipped with multiple heads and tails for multiple low-level vision tasks including deraining. Liang et al. \cite{liang2021swinir} propose a Swin Transformer-based model by combining CNN and transformer and achieves superior performance while maintaining computational efficiency. Wang et al. \cite{wang2021uformer} constructs a U-shape transformer based on Swin Transformer for image restoration. Apart from the U-shape architecture, Restormer~\cite{zamir2021restormer} utilizes the channel-wise attention and convolution and obtains better SID results.

\subsection{Deraining datasets}
Multiple datasets for SID have been proposed in previous works~\cite{fu2017removing,yang2017deep,zhang2019image,li2016rain,zhang2018density,hu2019depth,li2019heavy}, most of which are generated by first creating rain streaks using the photo-realistic rendering technique~\cite{garg2006photorealistic,garg2007vision} and then blending them with clean images. For example, Jiang et al. \cite{jiang2020multi} propose a large and diverse synthetic dataset containing 13k paired images gathered from multiple aforementioned datasets. As for real data, there are two paired datasets in this field as far as we know, named SPA-data~\cite{wang2019spatial} and RainDS~\cite{quan2021removing}, respectively.
When revisiting the rainy images in RainDS, we find that they are generated by using sprinklers in front of the camera to simulate the rainy scenes, leading to depth information loss and homogeneous rain streaks due to the less-varied rain-to-camera distance and the less-varied physical characteristics of the water from sprinklers.
We also identify the spatial misalignment issue in RainDS as indicated by the green rectangles in Figure~\ref{fig:dataset_glance} (a), which will cause erroneous loss that misleads the training process and result in inaccurate evaluation performance.
% The SPA-Data contains about 638k rain and clean image patches, which are semi-automatically generated from 170 real rain videos. RainDS contains 250 image pairs, which is constructed by spraying water in front of the background using sprinklers.
The rainy images in SPA-data are the frames from $170$ videos with static backgrounds, among which drizzling videos are dominating. Consequently, the small number of videos and sparse rain streaks in each image lead to the limited diversity of rain and background appearances as shown in Figure~\ref{fig:dataset_glance} (b), which cannot comprehensively reflect the complex and diverse scenes in real world with rain at various densities.
Besides, the images in SPA-data are small patches with low resolution as shown in Table~\ref{tab:dataset_comparison}.
Although these two datasets have advanced the research in this field, there are still some problems needed to be solved.
On the one hand, due to the domain gap between synthetic and real rainy images, SID models trained on existing synthetic datasets always do not generalize well on real-world rainy images. On the other hand, existing real-world datasets have limited diversity in rain streaks, backgrounds, and scenes, making it difficult to benchmark SID methods comprehensively. Therefore, it is necessary to establish a high-quality real-world dataset containing abundant scenes, diverse rain streaks, and various rain densities, and benchmark existing methods to show the overall progress in this area.

\section{The proposed datasets}
\subsection{The real rainy dataset: RealRain-1k}

\noindent \textbf{Data collection}
We first collect a large number of high-resolution rainy videos from diverse real-world scenes. Specifically, $972$ high-resolution rainy videos with static background are captured using a tripod to avoid the jittering artifacts and achieve strict spatial alignment. Moreover, we also collect $148$ online available videos and have $1,120$ videos in total, which cover various daylight and night outdoor scenes such as buildings, avenues, and parks. In addition, they have diverse rain streaks, different rain densities, and various backgrounds.

\begin{figure}[H]
    \centering
    \includegraphics[width=\linewidth]{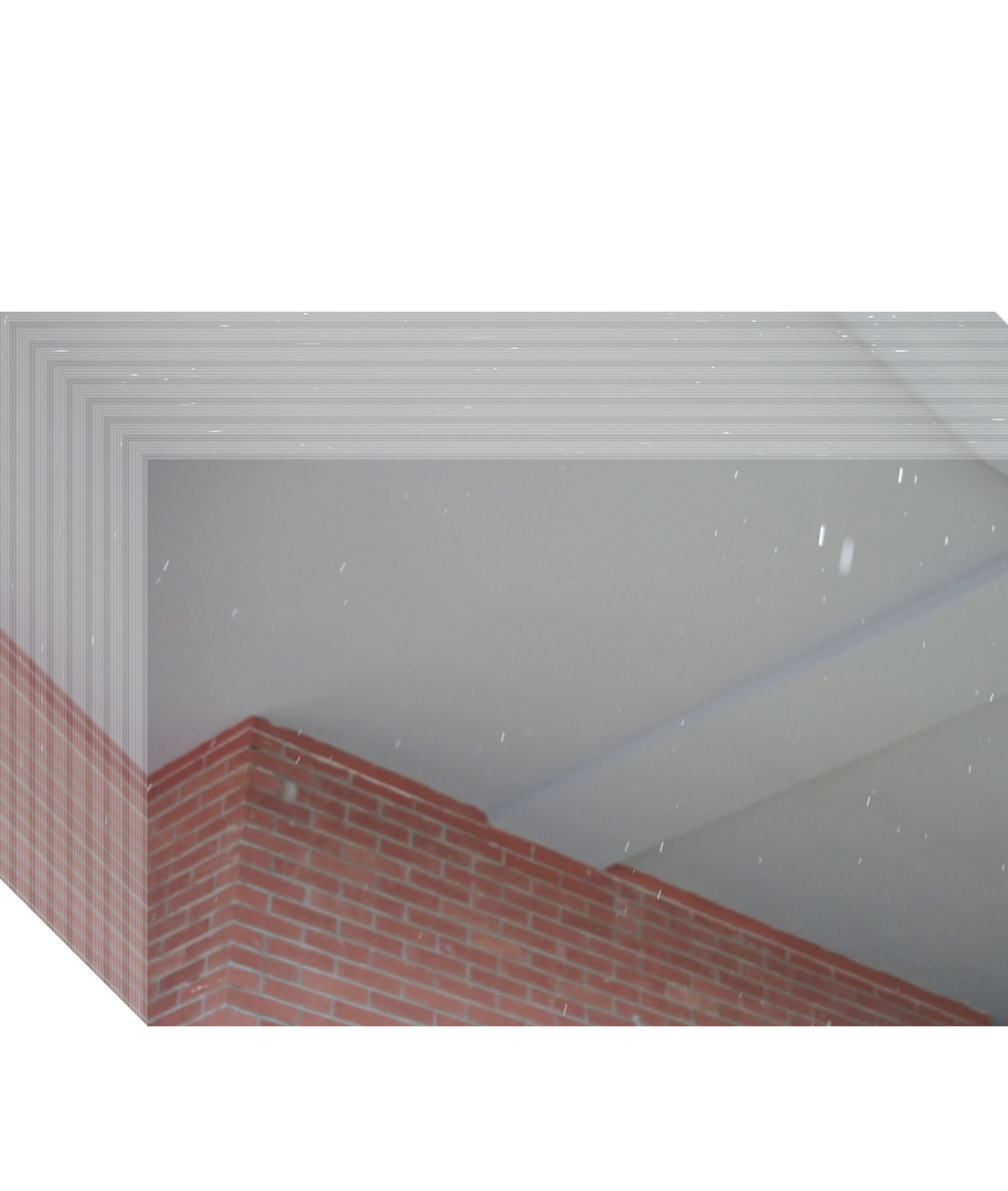}
    \caption{Overview of the generation pipeline of RealRain-1k.}
    \label{fig:realrain-1k_generation}
    % \vspace{-2 mm}
\end{figure}

\begin{figure}[htbp]
    \centering
    \includegraphics[width=\linewidth]{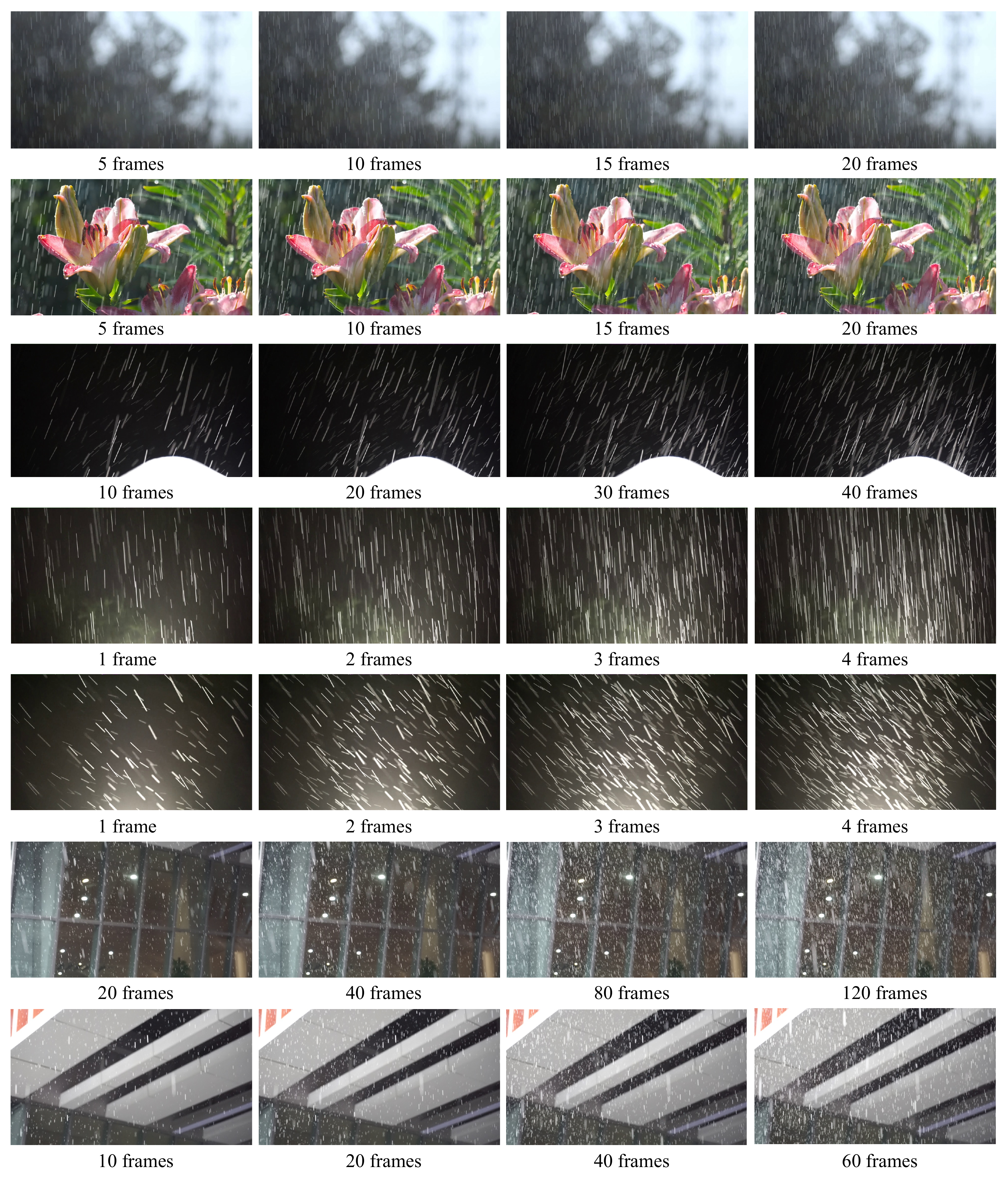}
    \caption{Some rainy images with different rain densities generated by the proposed method. The number below each image denotes the number of frames used for generating the rainy image based on the maximum filter.}
    \label{fig:max}
    % \vspace{-2 mm}
\end{figure}

\noindent \textbf{Triplet generation}
We notice that the rain in the real world significantly varies from light to heavy. To this end, we use a simple yet effective rain density-controllable filtering method to generate rainy and clean image pairs with different rain densities to mimic different rainy scenarios in the real world. The process is illustrated in Figure~\ref{fig:realrain-1k_generation}. Based on the observation that the pixel in a rainy image always has higher intensity if covered by rain streaks, we propose a simple maximum filtering method over the video clips to select and retain the bright pixels, \ie, pixels with rain. Specifically, given the pixel intensity $P_{(x,y)}^i$ at the position $(x,y)$ of the $i_{th}$ frame from a rainy video, the rainy images $R$ are generated as follows:
\begin{equation}
    R_{(x,y)}= \max\{ P_{(x,y)}^i | l \leqslant i \leqslant r \},
\label{eq:maximum}
\end{equation}
where $l$ and $r$ denote the selected start and end frame index in the video. Obviously, if a longer video clip is selected in Eq.~\eqref{eq:maximum} (\ie, a larger $|r-l|$) for rainy image generation, more pixels covered by rain streaks will be kept and thus the rain density increases, and \textit{vice versa}. In this way, we can mimic the real-world rainy images with various rain densities by simply adjusting the length of the selected video clip to control the rain density in the generated rainy images. Specifically, we generate rainy images with two levels of rain densities, \ie, light and heavy, respectively, where the clip length is determined manually for each video since our captured videos have different rain densities. Figure~\ref{fig:max} shows some examples with different rain densities.
%which results in 1,120 heavy-rain images and 1,120 light-rain images in RealRain-1k.

It is assumed that the pixel is infrequently covered by rain streaks in a video clip lasting for a long period, especially in the mild rainy conditions. Thereby, most pixels at the same spatial position of the consecutive frames may be clean pixel candidates. Based on this assumption, we propose a simple median filtering method to obtain a reference clean (background) image for each video. Specifically, let $B_{(x, y)}$ denotes the pixel value at the position $(x,y)$ in the generated clean image, it can be calculated as follows:
\begin{equation}
    B_{(x,y)}= {\rm median}\{ P_{(x,y)}^i | l \leqslant i \leqslant r \},
\label{eq:median}
\end{equation}
where $P_{(x, y)}^i$, $l$, and $r$ have the same meanings as Eq.~\eqref{eq:maximum}.

To obtain high-quality rainy and clean image pairs, we carefully check them during the generation process. Three rounds of cross-checking and re-generation (\eg, select the start frame index and adjust the length of the video clip) are carried out to ensure the quality of the generated image pairs, \ie, in terms of strict spatial alignment, clean images without rain streaks, various rain streak appearances, and diverse backgrounds. In total, we generate 1,120 image triplets (clean, light rain, heavy rain) respectively for two versions of our RealRain-1k dataset, \ie, RealRain-1k-L and RealRain-1k-H respectively.

\begin{figure*}[!t]
    \centering
    \includegraphics[width=\linewidth]{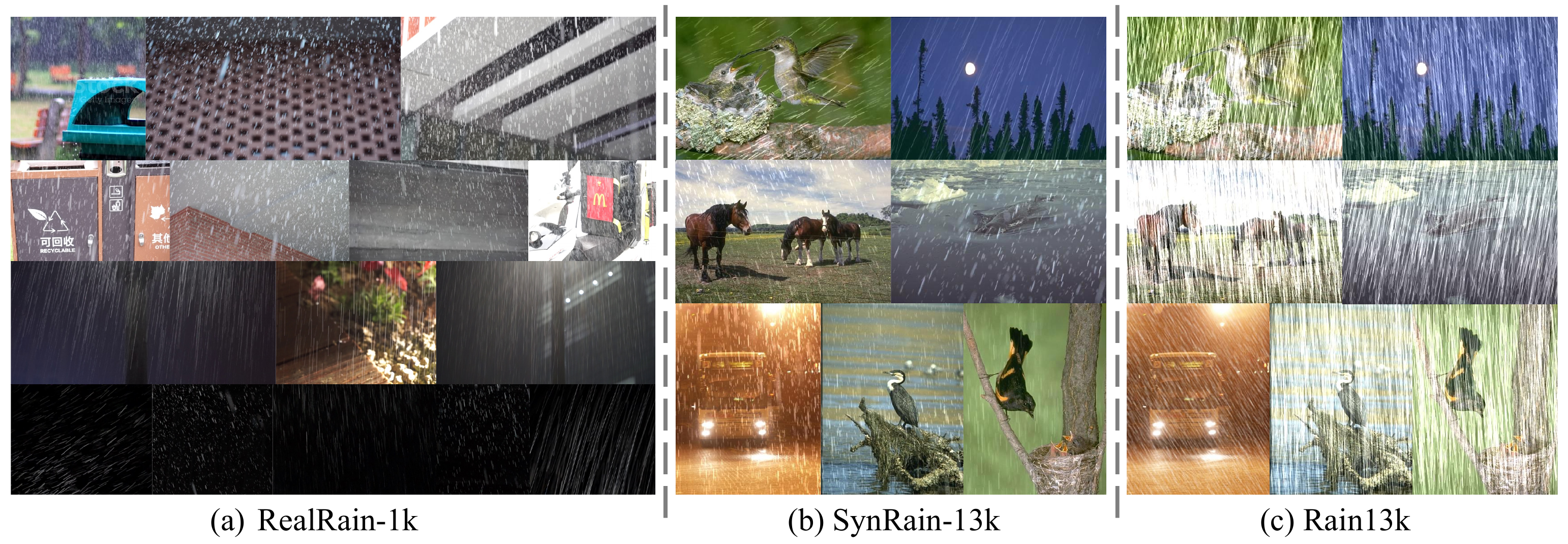}
    \caption{Some examples from our RealRain-1k, SynRain-13k, and Rain13k~\cite{jiang2020multi}. We show some real rain layers from our RealRain-1k in the bottom row of (a). We choose the rainy images with the same backgrounds from our SynRain-13k and Rain13k for a better comparison.}
    \label{fig:synthetic_datasets_glance}
    % \vspace{-2 mm}
\end{figure*}

\noindent \textbf{How to avoid the misalignment issue?} It's difficult to capture high-quality rainy and clean pairs with exact the same scene at the same time. We thus conduct the following strategies to guarantee the alignment quality in RealRain-1k: 1) Many images in our dataset consist of static objects (such as buildings, cars, etc.) that will not be affected when raindrops hit. 2) Thanks to the proposed rain density-controllable generation method, we don't have strict weather limitations to obtain the rainy videos with diverse rain densities. We instead choose the relatively normal rainy days to reduce the impact of the raindrops hitting the easy-to-move objects (flowers, leaves, etc.). Similarly, breezy days are selected to avoid wind disturbance. 3) We also carefully filter out some video clips with large disturbances (caused by wind, rain, and human disturbances) with many human efforts to ensure the quality of the dataset regarding alignment. Then each pair in our dataset undergoes repeated (at least three times) comparisons between rainy images and groundtruth. If disturbances are still unavoidable after filtering out the undesirable video clips, we choose to throw away the pairs from our dataset.
% Capturing the rainy images and clean images at two different timestamps is also undesirable as the illumination shifts may cause to unsatisfactory pairs, and it requires a lot of labor cost. Hence it's indeed difficult to collect high-quality rainy and clean pairs. 

We acknowledge that it is impossible to obtain completely aligned real rainy datasets, and thus we made many efforts to include strict human supervision in the loop to make the misalignment almost invisible to the naked eyes. Compared to current valuable real rainy datasets SPA-data and RainDS, the misalignment issue is relieved in RealRain-1k.
It is noted that we don't claim the alignment issue is completely resolved and don't mean to criticize existing real datasets. They have played an important role in promoting the development of real-world SID. We just hope RealRain-1k can make a positive supplement to the existing real rain datasets and our work can help to shed some light on the community of real-world SID. 

\begin{table*}[htbp]
\caption{The comparison of the proposed RealRain-1k and SynRain-13k datasets and existing ones. The number separated by the slashes in the second column denotes the number of image pairs in the training set, validation set, and test set in each dataset, respectively.}
  \centering
  \normalsize
  \resizebox{\linewidth}{!}{
    \begin{tabular}{c|c|c|c|c|c|c}
    \hline
    Dataset & Image pairs & Data source & Real & Average resolution & Features & Ratio of night scenes \\
    \hline
    RainDS~\cite{quan2021removing} & 150/0/98 & using sprinkler & \checkmark  & 1,296$\times$728 & \makecell[c]{not strictly aligned, \\homogenous rain streaks} & 0\% \\
    \hline
    SPA-data~\cite{wang2019spatial} & 638,492/0/1,000 & \makecell[c]{170 videos\\49\% videos are captured using mobile phone} & \checkmark  & 256$\times$256 & \makecell[c]{mostly light rain, \\homogeneous rain streaks} & 	7.53\% \\
    \hline
    Rain13k~\cite{jiang2020multi} & 13,711/0/4,298 & \makecell[c]{synthetic rain streaks\\ and background images} &  \ding{55}  &     482$\times$420  & \makecell[c]{large domain gap \\between real and synthetic}   & - \\
    \hline
    RealRain-1k & 784/112/224 & \makecell[c]{1,120 high-resolution videos\\ 87\% videos are captured using SONY digital camera} & \checkmark  & 1,512$\times$973 & \makecell[c]{\textbf{strictly aligned,} \\\textbf{diverse rain streaks,} \\\textbf{controllable rain density}} & 47.95\%\\
    \hline
    SynRain-13k & 13,711/198/1,200 & real rain layer + background &  \ding{55}  &  482$\times$420     & \makecell[c]{small domain gap \\between real and synthetic} & - \\
    \hline
    \end{tabular}%
    }
  \label{tab:dataset_comparison}%
\end{table*}%

\begin{figure}[t]
    \centering
    \includegraphics[width=1\linewidth]{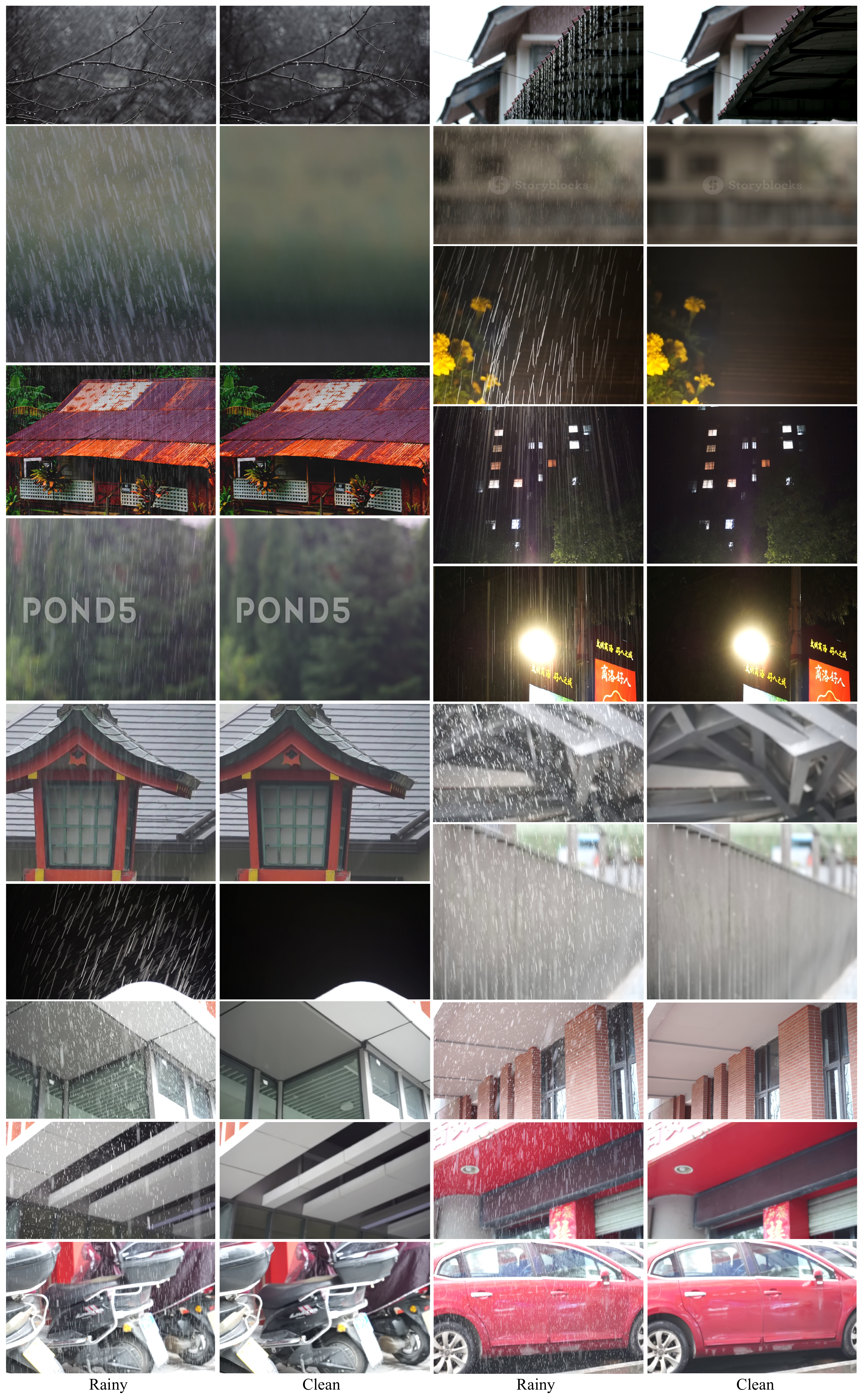}
    \caption{More examples from our RealRain-1k.}
    \label{fig:more_RealRain-1k}
    \vspace{1 mm}
\end{figure}

\noindent \textbf{Data organization and statistics} The 1,120 image pairs in RealRain-1k-L (and RealRain-1k-H) are further split into three disjoint subsets, \ie, training set, validation set, and test set, at the ratio of 7:1:2 respectively. Compared with SPA-data and RainDS, our RealRain-1k contains higher resolution images, \ie, an average resolution of 1,512$\times$973. Some examples are shown in Figure~\ref{fig:synthetic_datasets_glance} (a) and Figure~\ref{fig:more_RealRain-1k}. The statistics of the proposed RealRain-1k and previous representative datasets are summarized in Table~\ref{tab:dataset_comparison}.

\subsection{The synthetic rainy dataset: SynRain-13k}
\begin{figure}[htbp]
    \centering
    \includegraphics[width=0.95\linewidth]{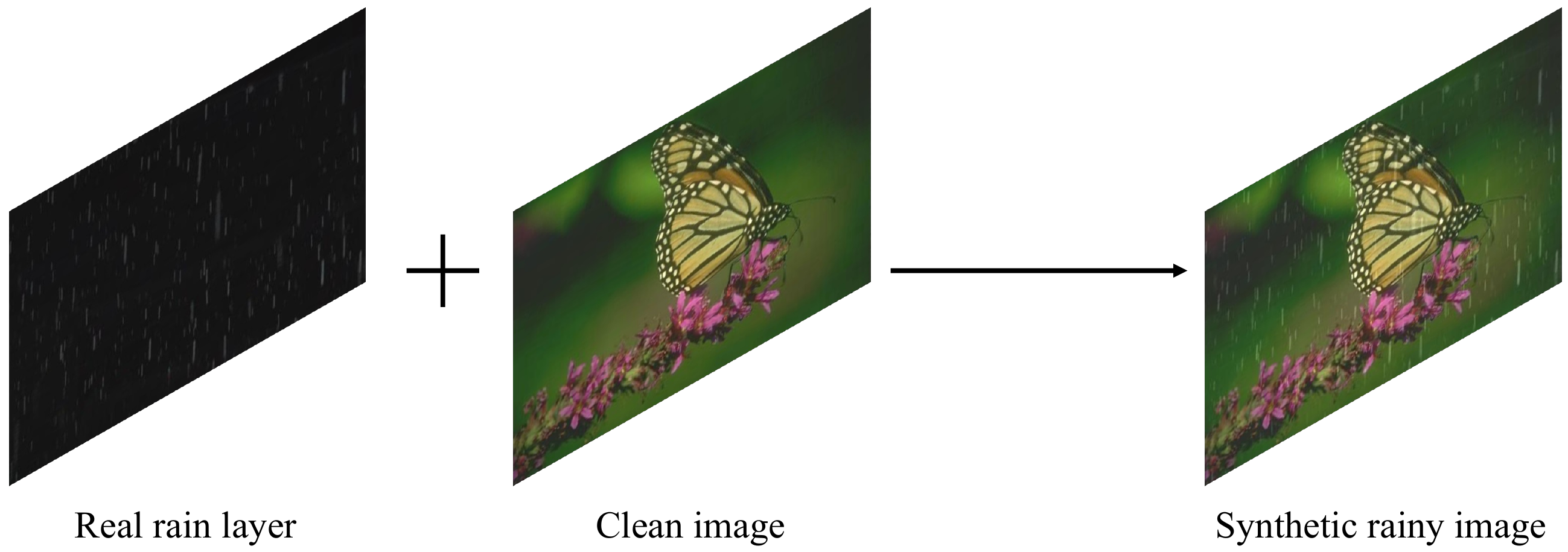}
    \caption{Overview of the generation pipeline of SynRain-13k.}
    \label{fig:SynRain-13k_generation}
    % \vspace{-2 mm}
\end{figure}

% \noindent{}
As a byproduct, we can extract the real rain layers from the rainy images in the training set of RealRain-1k-L after obtaining the rainy and clean image pairs. As shown in Figure~\ref{fig:SynRain-13k_generation}, it can be achieved by reversing the linear superimposition model as follows:
\begin{equation}
    L = R - B,
    \label{eq:superimposition}
\end{equation}
where $R$ and $B$ represent the rainy and clean images, respectively. After double-checking the appearances and density of the extracted rain streaks, we manually select 440 clean rain streak layers finally. Some examples are shown in the bottom row of Figure~\ref{fig:synthetic_datasets_glance} (a). Then, we randomly paste them on the clean images from Rain13k~\cite{jiang2020multi} following the linear superimposition model and the process is illustrated in Figure~\ref{fig:SynRain-13k_generation}. In this way, we generate a new large-scale synthetic rainy dataset containing 13k rainy images, \ie, SynRain-13k, by keeping the same dataset volume as Rain13k. Compared with previous synthetic datasets that are generated based on simulated rain streaks~\cite{yang2017deep,liu2021unpaired}, we directly use the real rain layers, thereby reducing the domain gap between the synthetic and real images, which is never explored before. Some examples are shown in Figure~\ref{fig:synthetic_datasets_glance} (b). As demonstrated in the experiment parts, SID models trained on our SynRain-13k show better generalization performance on real rainy images.

\section{Experiments}
Based on our proposed real and synthetic datasets RealRain-1k and SynRain-13k, we benchmark thirteen representative SID methods including two traditional ones \cite{luo2015removing,li2016rain} and eleven state-of-the-art deep learning-based ones \cite{zamir2021multi,zamir2021restormer,chen2021hinet,yi2021structure,deng2020detail,jiang2020multi,wang2019spatial,li2018recurrent,fu2019lightweight,fu2017clearing,fu2017removing}. 

Specifically, we conduct the experiments on three tracks, \ie, (1) supervised learning (SL) track, (2) domain generalization (DG) track, and (3) cross-domain transfer learning (TL) track. The detailed settings of each track will be presented later. We use the common PSNR and SSIM~\cite{wang2004image} metrics to measure the performance of different methods. In addition, we present some visual results for subjective evaluation.

\subsection{Supervised learning track}
\label{sec:Supervised learning track}
In this track, we compare different methods by training and testing them on our RealRain-1k-L and RealRain-1k-H, respectively. For each method, we use the original implementation code provided by the authors and adopt the same training settings including augmentation, batch size, learning rate, optimizer, etc. 
% Our baseline model follows Restormer~\cite{zamir2021restormer} for a fair comparison. 
We follow the common practice for all experiments, \ie, all the methods are trained on the training set, validated on the validation set, and evaluated on the test set.

\begin{table*}[htbp]
%\vspace{-1\baselineskip}
\caption{Quantitative results of different SID methods on RealRain-1k-L and RealRain-1k-H. 
%We use bold and underline to indicate the best and suboptimal performance, respectively.
}
  \centering
  \scriptsize
  \resizebox{\linewidth}{!}{
    \begin{tabular}{c|c|cc|cc|c|c}
    \toprule
     \multirow{3}[1]{*}{Category}     &   \multirow{2}[1]{*}{Method}    & \multicolumn{2}{c|}{RealRain-1k-H} & \multicolumn{2}{c|}{RealRain-1k-L} & Parameters & FLOPs \\
\cline{3-6}          &       & PSNR (dB) & SSIM  & PSNR (dB) & SSIM  & (M) & (G)\\
\cline{2-8} & Rainy images & 22.27 & 0.7657 & 25.93 & 0.8651 &  -  & - \\
    \hline
    \multirow{2}[2]{*}{Traditional} & DSC \cite{luo2015removing}  & 24.48&	0.7857	&27.76&	0.8750&	   -    &  - \\
          & GMM \cite{li2016rain}   & 24.93 & 0.8712 & 28.87 & 0.9259 &    -   & - \\
    \hline
    \multirow{10}[2]{*}{CNNs} & DDN \cite{fu2017removing} & 29.17 & 0.8783 & 31.18 & 0.9172 & 0.06  & 7.41 \\
          & DerainNet \cite{fu2017clearing} & 22.88 & 0.8886 & 27.09 & 0.9250 & 0.75  & 88.29 \\
          & RESCAN \cite{li2018recurrent} & 29.39 & 0.8914 & 31.33 & 0.9261 & 0.15  & 32.32 \\
          & LPNet \cite{fu2019lightweight} & 28.62 & 0.9424 & 32.47 & 0.9596 & 0.03 & 3.57 \\
          & SPANet \cite{wang2019spatial} & 25.76 & 0.9095 & 30.43 & 0.9470 & 0.28  & 36.25 \\
          & DRDNet \cite{deng2020detail} & 30.78 & 0.9308 & 31.94 & 0.9426 & 5.23  & 689.84 \\
          & MSPFN \cite{jiang2020multi} & 25.08 & 0.8045 & 35.51 & 0.9668 & 21.00  & 708.44 \\
          & SPDNet \cite{yi2021structure} & 26.63 & 0.8498 & 28.98 & 0.9005 & 3.32  & 96.61 \\
          & MPRNet \cite{zamir2021multi} & 34.74 & 0.9635 & 36.29 & 0.9721 & 3.64  &141.45  \\
          & HINet \cite{chen2021hinet}      &   40.82    & 0.9830 &41.98       &  0.9869     &     88.67  & 170.71 \\
    \hline
    Transformer & Restormer\cite{zamir2021restormer} & 39.57  & 0.9812  &   40.90  & 0.9849  &   26.10    &140.99  \\
    \bottomrule
    \end{tabular}%
    }
   \label{tab:supervised}%
\end{table*}%

\textbf{Main results}
The results as well as model size and FLOPs are summarized in Table \ref{tab:supervised}. As can be seen, the deep learning-based method of both architectures, \ie, CNNs and transformers, significantly outperform the traditional prior-based methods on both RealRain-1k-H and RealRain-1k-L. The results validate the superiority of the deep neural networks, owing to their strong representation ability and the data-driven learning paradigm. When comparing CNN-based and the advanced transformer-based methods, Restormer outperforms MSPFN by a large margin, although they have the similar model size. Comparing the results of each method on RealRain-1k-H and RealRain-1k-L, we find that the performance of all the methods drops significantly when dealing with heavy rainy images. It shows that the proposed RealRain-1k-H and RealRain-1k-L have different levels of difficulties as implied by their names. It is noteworthy that our datasets can be easily extended to cover different levels of rain densities and it is interesting to investigate how different methods perform on rainy images with different rain densities, which we leave as our future work. Among all the methods, HINet achieves the best performance of 40.82 dB and 41.98 dB on RealRain-1k-H and RealRain-1k-L, respectively. However, it is also the largest model that has 88M parameters and high computational cost, \ie, about 170G FLOPs. Therefore, it also matters to develop efficient SID methods that can achieve a better Pareto front between the deraining performance and model complexity.

% Thanks to the divide-and-conquer strategy, our baseline model can adaptively process the channel-wise and spatial-wise information to recover images from rain streaks, thus obtaining better performance than Restormer with minimal cost and help to set a Pareto efficiency baseline.

%\subsubsection{Convergence comparison}
\textbf{Convergence comparison}
%\textbf{Setting}
We select five representative methods to investigate their convergence performance by training them on RealRain-1k-L and RealRain-1k-H with different epochs, including 200, 500, 1000, and 2,000 epochs, respectively. For methods like Restormer that splits the training pipeline into several stages, we determine the training epochs in each stage proportionally.

%\textbf{Results and analysis} 
The results are shown in Table~\ref{tab:convergence}. As can be seen, except the smallest model, \ie, DDN whose performance fluctuates in different settings, all other methods benefit from a longer training schedule. When the model size increases, the model tends to have faster convergence speed owing to the stronger representation ability. For example, HINet trained with 200 and 500 epochs obtains comparable performance to MPRNet trained with 500 and 1000 epochs, respectively. %Besides, the architecture design also matters for good performance and fast convergence speed, \eg, the transformer-based Restormer with much fewer parameters performs comparably to HINet on RealRain-1k-H.
An interesting finding is that the superiority of HINet over MPRNet diminishes with the increase of training epochs, where HINet is 20$\times$ larger than MPRNet. It implies that big models may have better data efficiency and it also matters to investigate the performance of different methods under the few-shot learning paradigm. 
\begin{table}[!htbp]
%\vspace{-0.5\baselineskip}
\caption{Convergence results on RealRain-1k.}
  \centering
  \Huge
  \resizebox{\linewidth}{!}{
    \begin{tabular}{l|cccc|cccc|c}
    \toprule
    & \multicolumn{4}{c|}{RealRain-1k-H} & \multicolumn{4}{c|}{RealRain-1k-L} & Parameters  \\
    \cline{1-9}
    Training epochs & 200   & 500   & 1000  & 2000  & 200   & 500   & 1000  & 2000 & (M) \\
    \hline
    DDN \cite{fu2017removing}  & 29.00    & 29.42 & 27.37 & 28.64 & 31.22 & 32.13 & 30.23 & 32.88 & 0.06 \\
    RESCAN \cite{li2018recurrent}& 27.99 & 29.98 & 30.93 & 31.21 & 31.42 & 31.96 & 33.33 & 33.90 & 0.15 \\
    MPRNet \cite{zamir2021multi}& 32.80  & 35.90 & 37.63 & 39.68 & 34.60  & 37.84 & 39.64 & 41.34 & 3.64 \\
    HINet \cite{chen2021hinet}& 35.55 & 37.53 & 39.05 & 40.20  & 37.52 & 39.45 & 40.70  & 41.55 & 88.67 \\
    Restormer \cite{zamir2021restormer}& 35.33 & 37.63 & 38.94 & 40.39 & 36.88 & 39.12 & 40.34 & 41.40  & 26.10 \\
    \bottomrule
    \end{tabular}
    }
    %\vspace{-2 mm}
  \label{tab:convergence}
\end{table}
%This implies that current large SID models may not converge to a significantly better rain-removal performance in complex real-world scenarios with rainy images of diverse rain streaks and backgrounds, thus implying a worth-to-explore problem of how to design a large model for better Pareto efficiency.
\subsection{Domain generalization track}
\label{sec: Domain generalization track}
In this track, we quantitatively show the generalization performance of different methods, which are trained on a synthetic or real dataset and tested on an unseen real dataset. It is a common scenario when applying the developed SID methods in real-world applications, but has not been well studied. The experiments are conducted at two settings, \ie, synthetic-to-real DG and real-to-real DG.

%\subsubsection{Synthetic-to-real DG}
\textbf{Synthetic-to-real DG}
%\textbf{Setting} 
We benchmark 11 deep learning-based methods based on the synthetic datasets, \ie, Rain13k~\cite{jiang2020multi} and our SynRain-13k, and real datasets, \ie, SPA-data~\cite{wang2019spatial}, RainDS~\cite{quan2021removing}, and our RealRain-1k-L and RealRain-1k-H. All the methods are trained on the training set of these synthetic datasets and evaluated on the test sets of these real datasets, following the original settings in their papers. We use the models trained on Rain13k from their official repositories if provided.

\begin{figure*}[!htbp]
    \centering
    \includegraphics[width=0.83\linewidth]{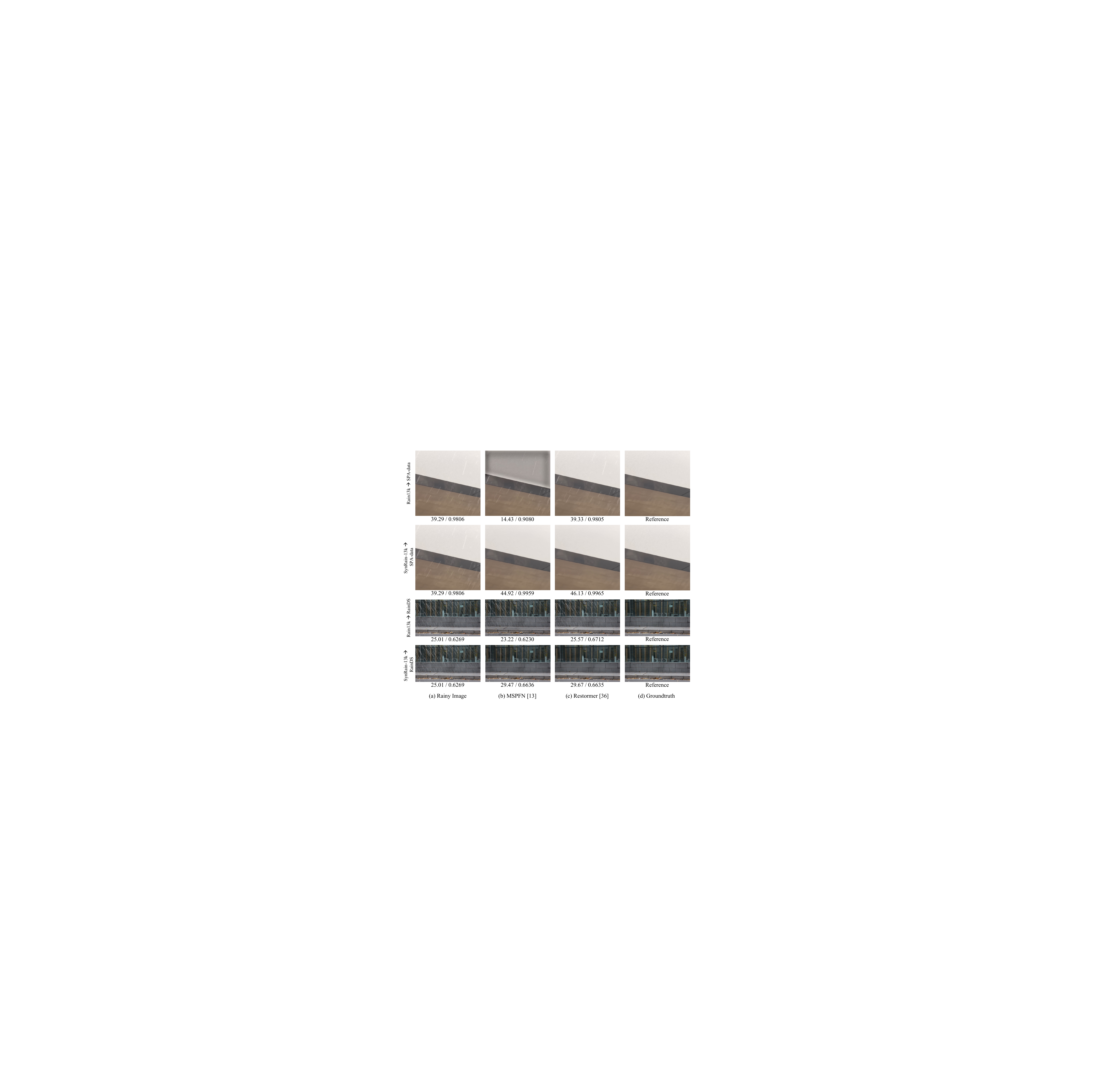}
    \caption{Visual comparison of three representative methods MSPFN \cite{jiang2020multi}, MPRNet \cite{zamir2021multi} and Restormer \cite{zamir2021restormer} trained on the Rain13k and our SynRain-13k respectively.}
    \label{fig:syn to real visual results}
    % \vspace{-2 mm}
\end{figure*}

The results are summarized in Table \ref{tab:syn-to-real-Dg}, Figure~\ref{fig:syn to real visual results} show some visual examples. As can be seen, when trained on SynRain-13k and tested on both SPA-data and RainDS, all the methods have better generalization performance than those trained on Rain13k. The results demonstrate the value of the proposed SynRain-13k which is generated using the real rain layers and has a smaller synthetic to real domain gap. 
%It is noted that our synthetic dataset generalizes better to our real rainy datasets, owing to the similar distribution of the rain streaks. 
Besides, Restormer trained on SynRain-13k achieves the best generalization performance in all scenarios, validating the good generalization ability of transformer as studied in \cite{zhang2021delving}.

We also show the generalization performance of different methods on the test sets of our RealRain-1k-H and RealRain-1k-L. As shown in Table \ref{tab:syn to RealRain-1k}, the models trained on our SynRain-13k generalize better on RealRain-1k, which is consistent with the results in Table \ref{tab:syn-to-real-Dg}, validating the value of the proposed SynRain-13k.
\begin{table*}[htbp]
\caption{Quantitative results of representative SID methods on the synthetic-to-real DG track.}
  \centering
  \scriptsize
  \resizebox{\linewidth}{!}{
    \begin{tabular}{c|cc|cc|cc|cc}
    \toprule
          & \multicolumn{2}{c|}{Rain13k $\rightarrow$ SPA-data} & \multicolumn{2}{c|}{SynRain-13k $\rightarrow$ SPA-data} & \multicolumn{2}{c|}{Rain13k $\rightarrow$ RainDS} & \multicolumn{2}{c}{SynRain-13k $\rightarrow$ RainDS} \\
          & PSNR (dB) & SSIM  & PSNR (dB) & SSIM  & PSNR (dB) & SSIM  & PSNR (dB) & SSIM \\
    \hline
    Rainy images & 32.63 & 0.9282 & 32.63 & 0.9282 & 21.80  & 0.6084 & 21.80  & 0.6084 \\
    \hline
    DDN \cite{fu2017removing}   & 30.13 & 0.9132 & 31.22 & 0.9288 & 22.31 & 0.5973 & 22.92 & 0.6319  \\
    DerainNet \cite{fu2017clearing} & 30.78 & 0.9118 & 30.83 & 0.9259 & 21.35 & 0.5914 & 22.45 & 0.6446 \\
    RESCAN \cite{li2018recurrent} & 32.04 & 0.9385 & 33.50  & 0.9510 & 23.18 & 0.6499 & 23.82 & 0.6629 \\
    LPNet \cite{fu2019lightweight} & 29.66 & 0.9201 & 31.10  & 0.9372 & 23.10  & 0.6429 & 23.29 & 0.6528 \\
    SPANet \cite{wang2019spatial} & 32.82 & 0.9337 & 33.81 & 0.9470 & 22.83 & 0.6289 & 23.21 & 0.6547 \\
    DRDNet \cite{deng2020detail} & 28.35 & 0.8943 & 31.56 & 0.9403 & 22.78 & 0.6164 & 23.40  & 0.6513 \\
    MSPFN \cite{jiang2020multi} & 22.46 & 0.8952 & 34.89 & 0.9548 & 17.33 & 0.5865 & 23.60  & 0.6628 \\
    SPDNet \cite{yi2021structure} & 32.44 & 0.9343 & 33.88 & 0.9531 & 23.41 & 0.6642 & 23.95 & 0.6706 \\
    MPRNet \cite{zamir2021multi} & 32.77 & 0.9396 & 35.08 & 0.9555 & 23.25 & 0.6563 & 23.82 & 0.6686 \\
    HINet \cite{chen2021hinet} & 32.35 & 0.9361 & 34.62 & 0.9521 & 23.50  & 0.6639 & 23.91 & 0.6683 \\
    Restormer \cite{zamir2021restormer} & 32.65 & 0.9358 & 35.48 & 0.9586 & 22.96 & 0.6480 & 24.03 & 0.6711 \\
    \bottomrule
    \end{tabular}%
    }
  \label{tab:syn-to-real-Dg}%
\end{table*}%
\begin{table*}[!htbp]

 \caption{Quantitative results of representative SID methods on the synthetic-to-real DG track.}
   \centering
   \scriptsize
   \resizebox{\linewidth}{!}{
     \begin{tabular}{c|cc|cc|cc|cc}
     \toprule
           & \multicolumn{2}{c|}{Rain13k $\rightarrow$ RealRain-1k-H} & \multicolumn{2}{c|}{SynRain-13k $\rightarrow$ RealRain-1k-H} & \multicolumn{2}{c|}{Rain13k $\rightarrow$ RealRain-1k-L} & \multicolumn{2}{c}{SynRain-13k $\rightarrow$ RealRain-1k-L} \\
           & PSNR (dB) & SSIM  & PSNR (dB) & SSIM  & PSNR (dB) & SSIM  & PSNR (dB) & SSIM \\
     \hline
     Rainy images & 22.27 & 0.7657 & 22.27 & 0.7657 & 25.93 & 0.8651 & 25.93 & 0.8651 \\\hline
     DDN \cite{fu2017removing}   & 27.00    & 0.8468 & 26.06 & 0.8270 & 29.94 & 0.9047 & 29.89 & 0.9016 \\
     DerainNet \cite{fu2017clearing} & 21.91 & 0.7420 & 23.50  & 0.8483 & 25.40  & 0.8474 & 27.29 & 0.9138 \\
     RESCAN \cite{li2018recurrent}& 25.32 & 0.8591 & 28.66 & 0.9057 & 28.98 & 0.9178 & 32.55 & 0.9460 \\
     LPNet \cite{fu2019lightweight}& 26.66 & 0.8824 & 27.54 & 0.8934 & 28.97 & 0.9207 & 31.16 & 0.9373 \\
     SPANet \cite{wang2019spatial}& 24.17 & 0.8304 & 24.96 & 0.8440 & 27.40  & 0.8959 & 30.33 & 0.9399 \\
     DRDNet \cite{deng2020detail}& 25.05 & 0.8131 & 27.42 & 0.8801 & 27.03 & 0.8634 & 31.27 & 0.9316 \\
     MSPFN \cite{jiang2020multi}& 17.92 & 0.7728 & 30.05 & 0.9399 & 19.50  & 0.8552 & 34.44 & 0.9678 \\
     SPDNet \cite{yi2021structure}& 24.63 & 0.8585 & 29.78 & 0.9272 & 27.94 & 0.9111 & 33.69 & 0.9590 \\
     MPRNet \cite{zamir2021multi}& 23.39 & 0.8346 & 30.32 & 0.9402 & 26.51 & 0.8949 & 34.40  & 0.9646 \\
     HINet \cite{chen2021hinet}&  23.82 & 0.8382 & 30.66 & 0.9390 & 26.56 & 0.8920 & 34.74 & 0.9651  \\
     Restormer \cite{zamir2021restormer}& 23.13 & 0.8152 & 31.89 & 0.9485 & 26.29 & 0.8825 & 35.90  & 0.9703 \\
     \bottomrule
     \end{tabular}
     }
   \label{tab:syn to RealRain-1k}
 \end{table*}

\begin{table*}[!h]
\caption{Quantitative results of representative SID methods on the real-to-real DG track.}
\vspace{+1.5\baselineskip}
  \centering
  \scriptsize
  \resizebox{\linewidth}{!}{
    \begin{tabular}{c|cc|cc|cc|cc}
    \toprule
          & \multicolumn{2}{c|}{RealRain-1k-L $\rightarrow$ SPA-data} & \multicolumn{2}{c|}{RealRain-1k-H $\rightarrow$ SPA-data} & \multicolumn{2}{c|}{SPA-data $\rightarrow$ RealRain-1k-L} & \multicolumn{2}{c}{SPA-data $\rightarrow$ RealRain-1k-H} \\
        %   \cline{2-9}
          & PSNR (dB) & SSIM  & PSNR (dB) & SSIM  & PSNR (dB) & SSIM  & PSNR (dB) & SSIM \\
    \hline
    Rainy images & 32.63 & 0.9282 & 32.63 & 0.9282 & 25.93 & 0.8651 & 22.27 & 0.7657 \\\hline
    MPRNet \cite{zamir2021multi} & 32.50  & 0.9482 & 30.70  & 0.9360 & 31.14 & 0.9519 & 27.96 & 0.9199 \\
    HINet \cite{chen2021hinet}& 35.37 & 0.9634 & 34.08 & 0.9575 & 32.45 & 0.9586 & 28.62 & 0.9225 \\
    Restormer \cite{zamir2021restormer}& 33.84 & 0.9590 & 32.93 & 0.9545 & 33.45 & 0.9687 & 28.69 & 0.9337 \\
    \bottomrule
    \end{tabular}%
    }
  \label{tab:real-to-real-DG}%
\end{table*}%

\textbf{Real-to-real DG}
%\textbf{Settings in real to real.} 
We benchmark three most representative SID methods in this track based on RealRain-1k-H, RealRain-1k-L, and SPA-data, whose rain density decreases in order. The models trained on RealRain-1k are collected from the experiments in Table \ref{tab:supervised}, while the models trained on SPA-data are collected from the official repositories or retrained following the same setting. 
% The results are given in Table~\ref{tab:real-to-real-DG}.

\begin{figure*}[htbp]
    \centering
    \includegraphics[width=\linewidth]{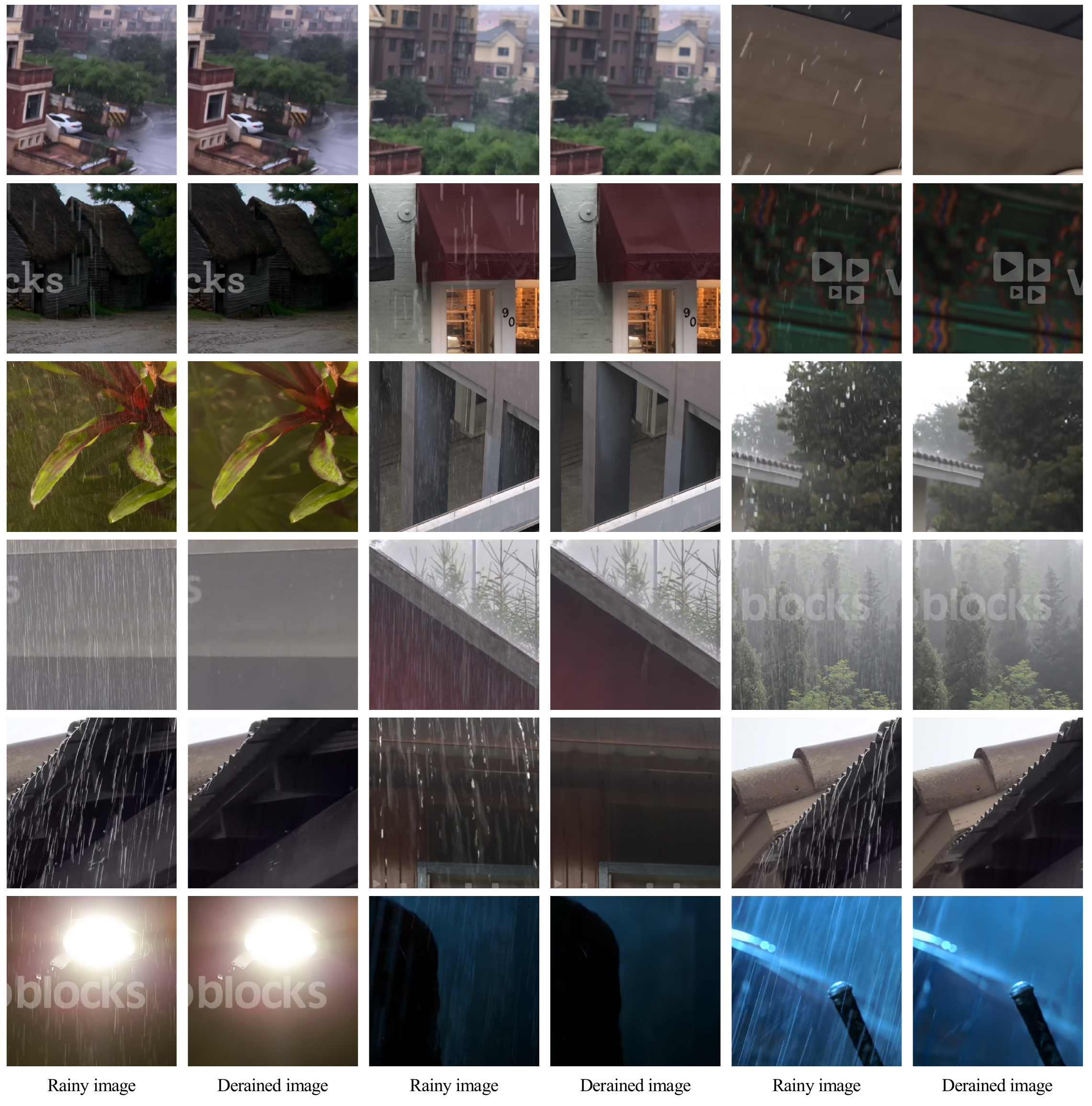}
    \caption{Some deraining results of HINet \cite{chen2021hinet} which is trained on our RealRain-1k-L and evaluated on SPA-data \cite{wang2019spatial}. We can see that the deraining results are fairly promising demonstrating that our datasets can generalize to various rainy scenarios.}
    \label{fig:real_to_real_visual_results_hinet}
    % \vspace{-2 mm}
\end{figure*}

The results are summarized in Table~\ref{tab:real-to-real-DG}. Comparing Table~\ref{tab:real-to-real-DG} and Table~\ref{tab:syn-to-real-Dg}, we can see that our RealRain-1k datasets can help methods generalize well to an unseen real domain. For example, both Restormer and HINet in the RealRain-1k-L$\rightarrow$SPA-data setting perform better than the counterparts in the Rain13k$\rightarrow$SPA-data setting. It is also interesting to find that the methods trained on our SynRain-13k outperform those trained on our RealRain-1k, owing to the real rain layers and large data volume of SynRain-13k. Besides, the experiment results also show that the generalization performance decreases when the difference of rain density in training and test sets becomes larger. For example, the methods in RealRain-1k-L$\rightarrow$SPA-data generalize better than those in RealRain-1k-H$\rightarrow$SPA-data. We also show some visual examples in Figure~\ref{fig:real_to_real_visual_results_hinet}. The promising deraining results further validate the value and effectiveness of RealRain-1k.
%We can see that the model trained on our real dataset can generalize to various real scenarios from sparse rain to heavy rain, day to night, rain streaks to raindrops.

To better reveal how and why the models behave differently when encountering the different rain density gap. We conduct several experiments to further support our claim. We split each test image into two non-overlapped regions according to the difference threshold between the rainy image and groundtruth, \ie, rainy region and rain-free region, where the rainy region denotes the area whose difference is above the threshold. We show the restoration performance on the two kinds of regions in the following Table~\ref{tab:rain_and_rain_free_region_1} and Table~\ref{tab:rain_and_rain_free_region_2}. It can be seen that the generalization performance from heavy rain to light rain is better in the rainy regions but affected by the unsatisfactory generalization ability in the rain-free regions. Specifically, when comparing RealRain-1k-H$\rightarrow$SPA-data and RealRain-1k-L$\rightarrow$SPA-data, the performance gain in rainy regions remains similar while there is a significant difference in the rain-free regions. This is because that the rain streaks cover more regions for those images with higher rain density. When handling the light rainy images where rain-free regions dominate, the model trained with RealRain-1k-H may remove rain-like structures in the rain-free regions, \eg, an unpleasant result of model overfitting. Therefore, it affects the overall gain brought by rain removal, leading to a relatively small PSNR improvement. In other word, it is caused by the limited generalization ability of deraining models due to domain gap between the training set and test set. The results in Table~\ref{tab:real-to-real-DG} also demonstrate this claim, where the performance gain of RealRain-1k-L$\rightarrow$SPA-data is larger than RealRain-1k-H$\rightarrow$SPA-data, and the models trained on SPA-data perform better on RealRain-1k-L than RealRain-1k-H, especially for Restormer. On the contrary, the model trained using light rainy images like SPA-data learns to recover only a small part of the images and tends to reserve most regions by regarding them as rain-free ones when dealing with the heavy rainy images. Consequently, the generalizing performance of all models from SPA-data to RealRain-1k-H reaches no more than 28.7dB, which is unsatisfactory because only part of rain streaks are removed. 
\begin{table*}[htbp]
  \centering
  \caption{The performance gain of rainy region and rain-free region when RealRain-1k-H/L generalize to SPA-data.}
  \scriptsize
   \resizebox{\linewidth}{!}{
    \begin{tabular}{c|cc|c|cc}
    \toprule
    RealRain-1k-H $\rightarrow$ SPA-data& \makecell[c]{Rainy region\\PSNR (dB)} & \makecell[c]{Rain-free region\\PSNR (dB)} & RealRain-1k-L $\rightarrow$ SPA-data& \makecell[c]{Rainy region\\PSNR (dB)} & \makecell[c]{Rain-free region\\PSNR (dB)}\\
    \hline
    Rainy images& 22.96 &37.34 &Rainy images &22.96 &37.34 \\\hline
    MPRNet\cite{zamir2021multi}	&26.95 &30.65 &MPRNet \cite{zamir2021multi} &26.83 & 33.31 \\
    Gain	&3.99 	&-6.68  &Gain	&3.87 	&-4.02\\
    \hline
    Restormer\cite{zamir2021restormer}	&27.59 &	31.60& Restormer \cite{zamir2021restormer}&	27.83&	32.71\\
    Gain&4.63 &-5.74 & Gain&	4.87 &	-4.62 \\
    \bottomrule
    \end{tabular}}
  \label{tab:rain_and_rain_free_region_1}%
\end{table*}%

\begin{table*}[htbp]
%\vspace{-1\baselineskip}
  \centering
  \caption{The performance gain of rainy region and rain-free region when SPA-data generalize to RealRain-1k-H/L.}
  \scriptsize
   \resizebox{\linewidth}{!}{
    \begin{tabular}{c|cc|c|cc}
    \toprule
    SPA-data $\rightarrow$ RealRain-1k-H  & \makecell[c]{Rainy region\\PSNR (dB)} & \makecell[c]{Rain-free region\\PSNR (dB)} & SPA-data $\rightarrow$ RealRain-1k-L& \makecell[c]{Rainy region\\PSNR (dB)} & \makecell[c]{Rain-free region\\PSNR (dB)}\\
    \hline
    Rainy images& 22.72 &	29.38  &Rainy images &23.72 &	31.15\\\hline
    MPRNet \cite{zamir2021multi}	&25.73&	31.78  &MPRNet \cite{zamir2021multi} &27.39 &	33.47  \\
    Gain	&3.01&	2.40  &Gain	&3.67 &	2.32 \\
    \hline
    Restormer \cite{zamir2021restormer}	&25.21&	31.23 & Restormer \cite{zamir2021restormer}&	27.88 &	33.33\\
    Gain&2.49 &	1.85 & Gain&4.16&	2.18 \\
    \bottomrule
  \end{tabular}}
  \label{tab:rain_and_rain_free_region_2}%
\end{table*}%
Our DG experiments reveal that the models behave differently when generalizing from heavy to light and light to heavy, \ie, models trained on light rain images can obtain larger performance gain when generalizing to heavy rain images, while the model generalizing from heavy to light tends to change the details of the rain-free region, resulting in relatively lower improvement. These findings have never been revealed by previous works, owing to the comparisons on the proposed rain density-controllable dataset. It confirms the value of our proposed dataset. Besides, it is also noteworthy that the proposed RealRain-1k-L and RealRain-1k-H are more challenging than the SPA-data since their rainy images have smaller PSNR values due to the larger rain density. As shown in Table~\ref{tab:real-to-real-DG} in the paper, the performance is far from saturation on the RealRain-1k-H dataset, calling for more research efforts on image deraining, especially for heavy rain images.

\textbf{Real-to-real DG with different training schedules} The results on the setting of real-to-real domain generalization with different training schedules are shown in Table \ref{tab:RealRain-1k-L->SPA} and Table \ref{tab:RealRain-1k-H->SPA}. As can be seen, the generalization performance on the SPA-data is getting better with the increase of training epochs on our RealRain-1k dataset. Again, the results confirm the value of our dataset for training a better model that generalizes well to real rainy images. It is noteworthy that the results of different methods in Table \ref{tab:RealRain-1k-L->SPA} and Table \ref{tab:RealRain-1k-H->SPA} are not consistent with those in Table \ref{tab:real-to-real-DG}, because we train them according to their default settings (i.e., with different training epochs for different methods) for the experiment in Table \ref{tab:real-to-real-DG}.

\begin{table*}[htbp]
  \centering
  \caption{RealRain-1k-L$\rightarrow$SPA-data domain generalization with different training schedules.}
  \scriptsize
   \resizebox{\linewidth}{!}{
    \begin{tabular}{c|cc|cc|cc|cc}
    \toprule
          & \multicolumn{8}{c}{RealRain-1k-L $\rightarrow$ SPA-data} \\\hline
    \multirow{2}[1]{*}{Training epochs}  & \multicolumn{2}{c|}{200} & \multicolumn{2}{c|}{500} & \multicolumn{2}{c|}{1000} & \multicolumn{2}{c}{2000} \\
          & PSNR (dB) & SSIM  & PSNR (dB) & SSIM  & PSNR (dB) & SSIM  & PSNR (dB) & SSIM \\\hline
    Rainy images & 32.63 & 0.9282 & 32.63 & 0.9282 & 32.63 & 0.9282 & 32.63 & 0.9282 \\\hline
    MPRNet \cite{zamir2021multi} & 30.94 & 0.9391 & 33.10  & 0.9512 & 33.63 & 0.9529 & 34.73 & 0.9601 \\
    HINet \cite{chen2021hinet}& 32.91 & 0.9489 & 34.26 & 0.9556 & 34.74 & 0.9593 & 35.14 & 0.9623 \\
    Restormer \cite{zamir2021restormer}& 32.75 & 0.9451 & 33.49 & 0.9545 & 33.65 & 0.9575 & 34.52 & 0.9623 \\
    \bottomrule
    \end{tabular}}
  \label{tab:RealRain-1k-L->SPA}%
\end{table*}%

\begin{table*}[htbp]
  \centering
  \caption{RealRain-1k-H$\rightarrow$SPA-data domain generalization with different training schedules.}
  \scriptsize
   \resizebox{\linewidth}{!}{
    \begin{tabular}{c|cc|cc|cc|cc}
    \toprule
          & \multicolumn{8}{c}{RealRain-1k-H $\rightarrow$ SPA-data} \\\hline
    \multirow{2}[1]{*}{Training epochs}  & \multicolumn{2}{c|}{200} & \multicolumn{2}{c|}{500} & \multicolumn{2}{c|}{1000} & \multicolumn{2}{c}{2000} \\
          & PSNR (dB) & SSIM  & PSNR (dB) & SSIM  & PSNR (dB) & SSIM  & PSNR (dB) & SSIM \\\hline
    Rainy images & 32.63 & 0.9282 & 32.63 & 0.9282 & 32.63 & 0.9282 & 32.63 & 0.9282 \\\hline
    MPRNet \cite{zamir2021multi}& 28.56 & 0.9175 & 31.80  & 0.9446 & 32.11 & 0.9455 & 34.01 & 0.9568 \\
    HINet \cite{chen2021hinet}& 30.44 & 0.9326 & 32.13 & 0.9455 & 33.37 & 0.9516 & 33.75 & 0.9555 \\
    Restormer \cite{zamir2021restormer}& 31.65 & 0.937 & 33.19 & 0.9508 & 33.14 & 0.9525 & 33.92 & 0.9590\\
    \bottomrule
    \end{tabular}}
  \label{tab:RealRain-1k-H->SPA}%
\end{table*}%

\subsection{Cross-domain transfer learning track}
\label{sec: Cross-domain transfer learning track}
In this track, we compare two representative methods, \ie, MPRNet \cite{zamir2021multi} and Restormer \cite{zamir2021restormer} on different settings. Compared to the synthetic-to-real DG setting, we collect the models trained on the synthetic datasets including Rain13k and SynRain-13k, and further fine-tune them on the training sets of SPA-data, RainDS, RealRain-1k-H, and RealRain-1k-L. The results are given in Table~\ref{tab:transfer}. Again, both methods pre-trained on our SynRain-13k outperform those pre-trained on Rain13k or without pre-training, showing the benefit of the real rain layers. And the results after fine-tuning are much better than those in the synthetic-to-real and real-to-real DG settings, which is reasonable since the training data from the downstream datasets are used. Note that the performances of Restormer \cite{zamir2021restormer} pre-trained on the synthetic datasets are inferior to the one 
without pre-training. Maybe it is because the backgrounds in the training set of SPA-data overlap with the ones in the test set. Therefore, Restormer could achieve a very good performance even without pre-training. 

\begin{table*}[htbp]
\vspace{-1\baselineskip}
  \centering
  %\small
  \caption{Results on the cross-domain transfer learning track.}
  \scriptsize
  \resizebox{\linewidth}{!}{
    \begin{tabular}{c|c|cc|cc|cc}
    \toprule
      \multirow{3}[0]{*}{Method}    &   \multirow{3}[0]{*}{Test set}    & \multicolumn{6}{c}{Pre-training dataset} \\
    \cline{3-8}
        & & \multicolumn{2}{c|}{N/A} & \multicolumn{2}{c|}{Rain13k} & \multicolumn{2}{c}{SynRain-13k}\\\cline{3-8}
          &       & PSNR (dB) & SSIM  & PSNR (dB) & SSIM  & PSNR (dB) & SSIM \\\hline
    \multirow{4}[0]{*}{MPRNet \cite{zamir2021multi}} & SPA-data & 39.73 & 0.9799 & 44.40  & 0.9881 & 44.91 & 0.9888 \\
          & RainDS & 23.59 & 0.6498 & 25.48 & 0.7101 & 25.65 & 0.7107 \\
          & RealRain-1k-H & 34.74 & 0.9635 & 38.48 & 0.9795 & 38.98 & 0.9807 \\
          & RealRain-1k-L & 33.47 & 0.9592 & 40.64 & 0.9844 & 41.35 & 0.9867 \\
          \hline
    \multirow{4}[0]{*}{Restormer \cite{zamir2021restormer}} & SPA-data & 46.87   & 0.9903 & 46.02 & 0.9904 & 46.14 & 0.9905 \\
          & RainDS & 25.10 & 0.6982 & 25.62 & 0.7153 & 25.79 & 0.7165 \\
          & RealRain-1k-H & 39.57 & 0.9812 & 41.63 & 0.9869 & 41.88 & 0.9872 \\
          & RealRain-1k-L & 40.90  & 0.9849 & 42.76 & 0.9890 & 43.01 & 0.9895 \\
          \bottomrule
    \end{tabular}%
    }
  \label{tab:transfer}%
\end{table*}%

\subsection{Model complexity}
The model complexity of different methods is shown in Table~\ref{tab:model complexity}, including model parameters, FLOPs, GPU memory footprint, and inference latency. From the table, we can see that the model size, the latency and FLOPs increase in recent years, probably due to the advance of GPUs and the trend of pursuing better performance. For example, MPRNet needs about 3s to process an image, while HINet has 88M parameters, limiting their usage in practical applications. Therefore, it deserves more research efforts to achieve a better balance between deraining performance and model complexity.
\begin{table}[H]
%\vspace{-1\baselineskip}
\caption{Model complexity of representative SID methods. The FLOPs are calculated given a test image of 256$\times$256 resolution. The GPU Memory and Latency are calculated when processing an image of 512$\times$512 resolution. All results are obtained on the same GTX 1080 Ti GPU.}
  \Huge
  \centering
  \resizebox{\linewidth}{!}{
    \begin{tabular}{c|c|c|c|c|c}
    \toprule
    Method& Venue& Params. (M) & FLOPs (G) & Memory (G) & Latency (ms) \\
    \hline
    DDN \cite{fu2017removing}  &CVPR'17 & 0.058 & 7.407 & 3.40   & 24.28 \\
    DerainNet \cite{fu2017clearing}& TIP'17& 0.750  & 88.287 & 4.40   & 64.00 \\
    RESCAN \cite{li2018recurrent} &ECCV'18 &0.150 & 32.320 & 1.10   & 103.83 \\
    LPNet \cite{fu2019lightweight} & TNNLS'19&0.027 & 3.566 & 8.10   & 471.00 \\
    SPANet \cite{wang2019spatial}&CVPR'19 &0.284 & 36.250 & 1.08  & 1103.20 \\
    DRDNet \cite{deng2020detail}& CVPR'20&5.230  & 689.840 & 10.70  & 888.39 \\
    MSPFN \cite{jiang2020multi}&CVPR'20& 21.000 & 708.437 & 10.70  & 656.20 \\
    SPDNet \cite{yi2021structure}&ICCV'21& 3.320  & 96.610 & 1.89  & 269.66 \\
    MPRNet \cite{zamir2021multi}& CVPR'21&3.640  & 141.450 & 1.67  & 2901.00 \\
    HINet \cite{chen2021hinet}& CVPRW'21& 88.666 & 170.714 & 1.97  & 826.86 \\
    Restormer \cite{zamir2021restormer}&CVPR'22& 26.097 & 140.990 & 3.98  & 1684.93 \\
    \bottomrule
    \end{tabular}%
    }
  \label{tab:model complexity}%
\end{table}%

\section{Conclusion and discussion}
\label{sec:conclusion}
In this paper, we establish a high-quality real rainy dataset RealRain-1k using a simple but effective rain density-controllable filtering
method and a large-scale synthetic dataset SynRain-13k based on real rain layers. Based on them, we benchmark more than 10 deraining methods comprehensively on three tracks. The results demonstrate the value of the proposed datasets, the different representation abilities of CNNs- or transformers-based methods, and the generalization performance under different learning paradigms with different datasets. We hope RealRain-1k and SynRain-13k can serve as test beds to facilitate future research in this area.

\textbf{Limitation and discussion}
Our real dataset only has 1k images in the current version, which can be scaled up by collecting more videos. Meanwhile, more rainy images of different rain densities can be generated to form a series of datasets, each of which has a specific level of rain density.

\textbf{Social impact.} RealRain-1k and SynRain-13k can benefit the study of image deraining that is of great practical significance. Attention should be drawn to the generalization ability of models trained on our datasets, which may not be good in unseen scenes with extremely different rain streaks.

\bibliographystyle{ieee}
\bibliography{egbib}

\end{document}